\documentclass[10pt,twocolumn,letterpaper]{article}

\usepackage[margin=0.75in]{geometry}
\usepackage{times}
\usepackage{graphicx}
\usepackage{subcaption}
\usepackage{hyperref}

\usepackage{booktabs}
\usepackage{amssymb, amsmath}
\usepackage{multirow}
\usepackage{tabularx}

\title{Closing the Alignment-Maturity Gap in Federated Prototype Learning\thanks{
Work funded by Project PID2023-147404OB-I00 (MICIU/AEI/10.13039/501100011033; ERDF/EU; ESF+/EU), Horizon Europe (GA 101070381), the Ministry for Digital Transformation and Civil Service and Next-GenerationEU/PRTR (TSI-100925-2023-1) and by the Competitive Reference Groups of the Xunta de Galicia (grant ED431C 2026/38).  CITIC, as a member of the CIGUS Network, receives subsidies from the ``Xunta de Galicia" and from the ERDF Operational Programme Galicia 2021-2027 (Grant ED431G 2023/01).
}}

\author{
Mario Casado-Diez\textsuperscript{1}\thanks{mario.diez@udc.es}\quad
Alejandro Dopico-Castro\textsuperscript{1}\thanks{alejandro.dopico2@udc.es}\quad
Verónica Bolón-Canedo\textsuperscript{1}\thanks{veronica.bolon@udc.es}\quad
Bertha Guijarro-Berdiñas\textsuperscript{1}\thanks{berta.guijarro@udc.es}\\
\textsuperscript{1}CITIC, Universidade da Coruña, A Coruña, Spain
}

\date{}

\begin{document}

\maketitle

\begin{abstract}
Learning discriminative visual representations from distributed, heterogeneous data is a fundamental challenge in Federated Learning (FL). Prototype-based methods address statistical heterogeneity by sharing class-level representations across clients but create a distance-dependent gradient pressure that is particularly severe during early training rounds: alignment pressure applied to immature global prototypes, aggregated from noisy local representations, generates large gradients that suppress the emergence of local discriminative structure. The result is a poorly organized embedding space and degraded recognition performance, particularly under severe non-IID conditions. We propose \emph{FedSAP}, a framework that stabilises federated representation learning through two complementary mechanisms: a deterministic alignment curriculum that delays global alignment until local representations become stable and a geometry-driven proxy separation loss that enforces inter-class structure on the unit hypersphere using the existing prototype bank without introducing additional parameters or communication overhead. Together, these mechanisms produce compact, well-separated class clusters without altering the underlying communication protocol between federation’s participants. Experiments across three benchmarks and varying degrees of heterogeneity show gains of up to 4 percentage points over the prototype-based baselines evaluated, with improvements most pronounced under high heterogeneity. The representational nature of our framework further enables a straightforward extension to semi-supervised settings, where unlabelled data is incorporated with minimal modification, underscoring the generality of scheduled alignment as a design principle.

\end{abstract}

\section{Introduction}
\label{sec:intro}

Visual recognition systems deployed at scale increasingly operate over data that cannot be centralised, distributed across hospitals, mobile devices, or edge nodes~\cite{rieke2020future} with heterogeneous local distributions. Federated learning (FL)~\cite{mcmahan2017communication} offers a framework for training under these constraints, but statistical heterogeneity remains its central obstacle: when clients hold different class distributions, locally optimised models diverge, and naive parameter averaging produces a global model that does not adequately represent any client~\cite{li2020federated, zhao2018federated}.

Prototype-based methods reframe this problem at the level of representations rather than parameters. Instead of averaging model weights, approaches such as FedProto~\cite{tan2022fedproto} and FedNH~\cite{dai2023tackling} share class-level embedding statistics --prototypes-- that serve as semantic anchors, pulling local representations toward a common structure across clients. This substantially reduces communication cost and mitigates client drift but introduces a subtle problem: the quality of these anchors depends entirely on the maturity of the representations that produced them.

In early training rounds, local feature extractors have not yet learned discriminative structures, and the global prototypes aggregated from them are correspondingly noisy. Yet alignment pressure is applied from round one. As shown in Section~\ref{sec:methodology}, the gradient of the prototype alignment loss scales directly with the distance between local embeddings and global prototypes. In early rounds, this distance is large by construction --both because local representations are unrefined and because global prototypes are unreliable anchors. The result is a \emph{double penalty}: large alignment gradients dominate the cross-entropy signal precisely when local feature learning is most fragile, suppressing the emergence of discriminative structure and destabilising optimisation. This dilemma is exacerbated under severe heterogeneity, where prototype estimates are noisier and local distributions diverge more sharply. Existing methods, including FedProto, apply a static alignment weight that ignores this temporal dynamic entirely.

We propose FedSAP (\textbf{Fed}erated \textbf{S}cheduled and \textbf{A}daptive \textbf{P}rototypes), a framework that resolves this penalty treating federated prototype learning fundamentally as a \emph{representation learning} problem: the objective is not merely to align embeddings with global anchors but to shape an embedding space that is simultaneously discriminative, compact, and coherent across clients. FedSAP pursues this through two complementary mechanisms.

The first one is a deterministic linear warm-up schedule for the prototype alignment weight. Rather than applying full alignment pressure from the outset, FedSAP gradually increases this weight over a fixed number of rounds, giving local representations time to consolidate before being pulled toward global consensus. The schedule is fully determined at the start of the training, introduces no additional per-round communication, and provides a transparent. Unlike adaptive or data-dependent schedules, it requires no auxiliary signals and is robust to schedule shape.

The second one is a \emph{geometry-driven proxy separation} mechanism. Rather than introducing a separate set of learnable proxy parameters, which would increase local model complexity and require additional communication, FedSAP repurposes the global prototypes received from the server as a shared proxy bank. Embeddings and prototypes are projected onto the unit hypersphere, and a cosine-softmax objective enforces inter-class separation by pulling representations toward their target class prototype while repelling them from non-target classes~\cite{movshovitz2017no, kim2020proxy}.

Taken together, these mechanisms decouple the \emph{when} of global alignment from the \emph{how strongly} of inter-class separation, producing a learning trajectory that is stable in early rounds and increasingly discriminative in later ones-- without modifying the communication protocol or introducing meaningful overhead.

A natural consequence of grounding FedSAP in representation learning principles is that its core design is perfectly suited to environments beyond the fully supervised setting. In federated environments, labelled data is often scarce or unevenly distributed across clients, making semi-supervised learning a practically important regime. We demonstrate that FedSAP extends to this setting with minimal modification: the same scheduled alignment and proxy separation objectives apply directly, with unlabelled data incorporated through a straightforward pseudo-labelling strategy. We do not claim this as a primary contribution, but we include it as evidence that scheduled alignment is a general design principle rather than a solution tailored to a single protocol.

The main contributions of this work are the following:

\begin{itemize}
    \item We identify and motivate the \emph{alignment-maturity gap} in prototype-based FL, where global alignment is applied before local representations are mature.

    \item We propose FedSAP, which addresses this gap through a progressive alignment curriculum and proxy separation loss applied to the existing prototype bank, well-separated embeddings at identical communication cost to FedProto.

    \item We evaluate FedSAP across three benchmarks and at multiple levels of heterogeneity, demonstrate improvements over prototype-based baselines, and show that the framework extends naturally to semi-supervised federated settings.
\end{itemize}

The remainder of this paper is organised as follows. Section~\ref{sec:related_work} reviews related work in federated learning and prototype-based methods. Section~\ref{sec:methodology} details the FedSAP framework. Section~\ref{sec:results} presents experiments, comparisons, and ablations, including the semi-supervised extension. Finally, Section~\ref{sec:conclusions} concludes the paper.

\section{Related Work}
\label{sec:related_work}

\paragraph{Federated Learning under Statistical Heterogeneity}

Most FL methods mitigate statistical heterogeneity by exchanging and aggregating locally learned model parameters into a shared global model. Statistical heterogeneity across clients is the central challenge in FL, FedAvg~\cite{mcmahan2017communication} suffers from client drift when local data distributions diverge, motivating a range of corrective approaches. Regularisation-based methods~\cite{li2020federated,li2021ditto} constrain local updates to remain close to a shared global reference, trading personalisation for consistency. Adaptive interpolation schemes~\cite{deng2020adaptive} learn client-specific combinations of local and global models, while clustered approaches~\cite{sattler2020clustered} partition clients by similarity before aggregation. Despite their diversity, these approaches operate at the parameter level, and weight-space coupling is inherently fragile under severe non-IID conditions where local optima diverge substantially.

\paragraph{Representation-Level Sharing}

An alternative family of methods relocates global coupling from parameters to representations. FedRep~\cite{collins2021exploiting} separates a shared feature extractor from personalised prediction heads, limiting global alignment to the embedding space. FedPAC~\cite{xu2023personalized} extends this by aligning local feature centroids with global counterparts and combining classifier heads with theoretically grounded weights, showing that explicit representation alignment substantially improves generalisation under label distribution shift. Moon~\cite{Li_2021_CVPR} applies model-level contrastive regularisation, using the previous global model as a positive anchor. These methods share the premise that a well-structured embedding space transfers more reliably across clients than aligned model weights, a premise that motivates our own approach. 

\paragraph{Prototype-Based Federated Learning}

Prototype-based methods carry this idea to its logical limit by exchanging only class-level embedding statistics. FedProto~\cite{tan2022fedproto} aggregates local class prototypes at the server and uses them as semantic anchors for local training, substantially reducing communication while mitigating client drift. FedNH~\cite{dai2023tackling} addresses prototype instability by initialising prototypes uniformly on a hypersphere and infusing class semantics gradually, combating prototype collapse under class imbalance. FedMPS~\cite{yang2025fedmps} introduces multi-level prototype-based contrastive learning and soft label generation to improve robustness, at the cost of increased complexity and a modified communication protocol. FedPAC~\cite{xu2023personalized} further aligns local feature centroids with global class statistics, demonstrating that coupling representation alignment with classifier collaboration yields strong personalisation gains. Despite these advances, a shared limitation persists: alignment pressure between local and global prototypes is applied with a static weight from the first communication round, when prototypes are least reliable. The resulting gradient destabilisation and suppression of local discriminative structure have not been explicitly addressed through principled alignment scheduling--the gap that FedSAP targets.

\paragraph{Metric and Proxy-Based Representation Learning}

Deep metric learning offers principled objectives for shaping embedding spaces. Traditional metric learning methods optimise relative distances between individual samples, but often suffer from expensive sampling strategies and unstable optimisation~\cite{schroff2015facenet}. Proxy-based methods address this by representing each class with a fixed embedding vector: ProxyNCA~\cite{movshovitz2017no, teh2020proxynca++} reduces metric learning to a nearest-neighbour classification problem in the proxy space, while ProxyAnchor~\cite{kim2020proxy} treats each proxy as an anchor and exploits data-to-proxy relations for richer gradient signals. Because they optimise distances to global class-representative proxies rather than comparing pairs of individual samples directly, proxy-based approaches converge faster and are more robust to noisy labels. These properties are highly valuable in the federated setting where local data is scarce and heterogeneous. FedSAP adapts this principle by repurposing the global prototype bank as a proxy set, coupling the separation objective to the current state of global representations without additional parameters or communication.

\paragraph{Semi-Supervised Federated Learning.}
Semi-supervised federated learning (SSFL)~\cite{lin2021semifed, diao2022semifl} leverages unlabelled data on distributed clients to improve model quality when labels are scarce~\cite{jeong2020federated}. FedMatch introduced inter-client consistency regularisation with disjoint parameter learning for labelled and unlabelled data, establishing the canonical SSFL problem formulation. Prototype-based SSFL frameworks have since gained traction because lightweight class-representative reduce communication overhead while guiding pseudo-label assignment: ProtoFSSL~\cite{kim2022protofssl} shares prototypes across clients to produce pseudo-labels with consistency regularisation, significantly reducing communication cost relative to model-sharing SSFL methods. FedPPL~\cite{pan2025fedppl} leverages global prototypes to adaptively adjust pseudo-label confidence thresholds, mitigating confirmation bias in medical imaging settings. A common limitation of these frameworks is that labelled and unlabelled objectives are optimised simultaneously from the first round: under severe heterogeneity, unrefined early representations generate noisy pseudo-labels that propagate errors through aggregation. FedSAP addresses this by initiating the semi-supervised phase only after the supervised alignment curriculum has stabilised the embedding space, enabling reliable prototype-based pseudo-labelling from the outset.

\section{Method}
\label{sec:methodology}

\begin{figure*}[t]
    \centering
    \includegraphics[width=0.9\linewidth]{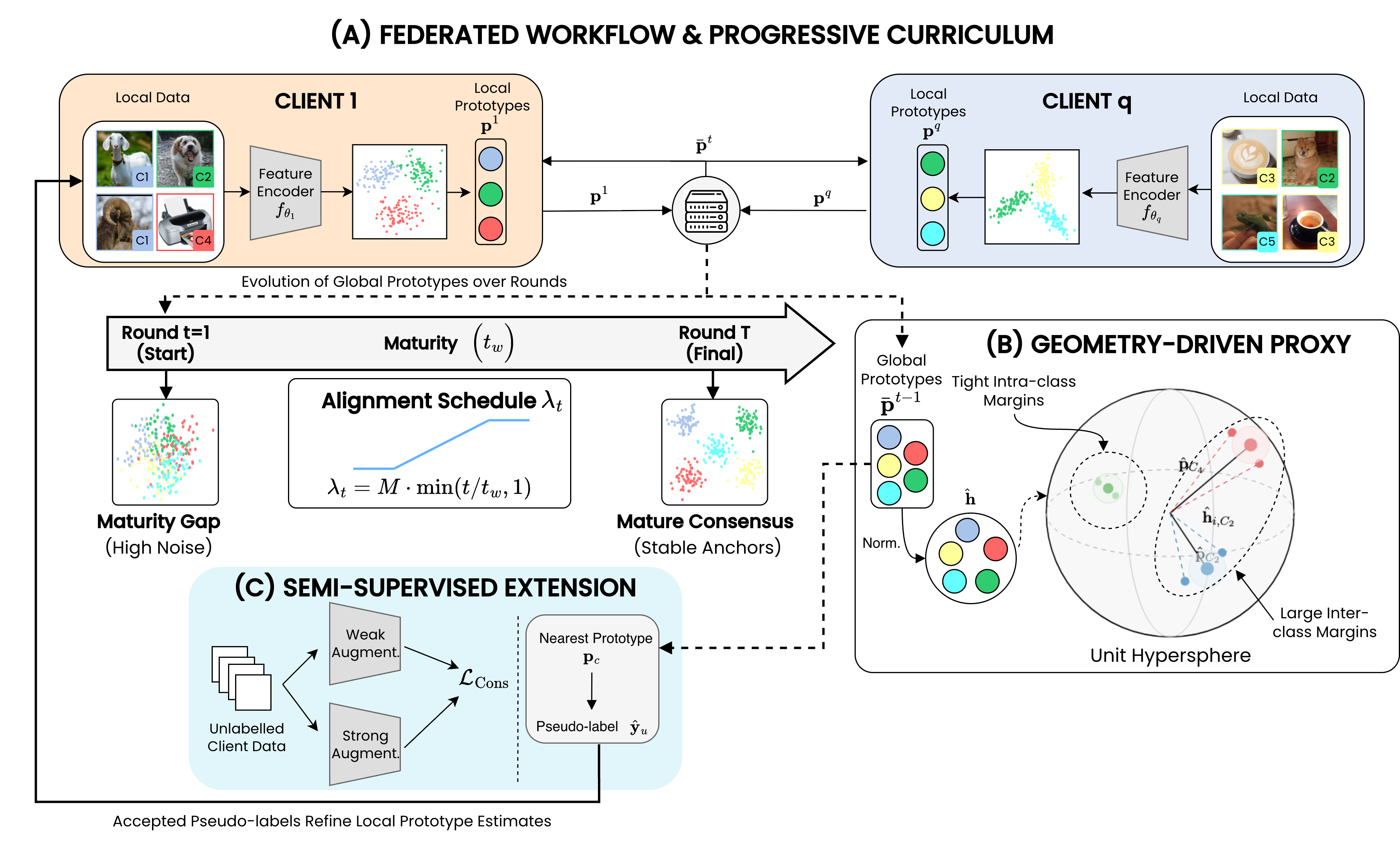}
    \caption{FedSAP architecture: a progressive alignment curriculum controls the onset of global prototype alignment, while geometry-driven proxy regularisation adapts inter-class separation to the current state of the embedding space.}
    \label{fig:method}
\end{figure*}

In this section, we present FedSAP (\textbf{Fed}erated \textbf{S}cheduled and \textbf{A}daptive \textbf{P}rototypes) a framework \ref{fig:method} designed to stabilise federated representation learning under statistical heterogeneity. The system is built around two coupled mechanisms: a progressive alignment curriculum that controls when global prototype pressure is introduced and a geometry-driven proxy regularisation that shapes inter-class separation using the same prototype bank exchanged with the server. Together, these mechanisms address the alignment-maturity gap identified below without modifying the communication protocol or introducing additional learnable parameters beyond the local feature extractor and classifier head.

\subsection{Problem Formulation and Workflow}

We consider a standard FL setup with a set of clients $\mathcal{Q} = \{1, \dots, Q\}$ coordinated by a central server over synchronised rounds $t = \{1, \dots, T\}$. Each client $q$ holds a local dataset $\mathcal{D}_q = \{(x_i, y_i)\}$ drawn from a client-specific distribution $\mathcal{P}_q$, with $\mathcal{P}_q \neq \mathcal{P}_{q'}$. Each client observes only a subset $\mathcal{C}_q \subset \mathcal{C}$ of the global class set, with $|\mathcal{C}_q| \ll |\mathcal{C}|$ of the global class set $\mathcal{C}$ under our partitioning protocol.

Each client maintains a feature extractor $f_{\theta_q}: \mathcal{X} \rightarrow \mathbb{R}^d$ and a classifier head $g_{\phi_q}$. At the end of each round $t$, client $q$ computes a local prototype for each observed class $c$ by averaging embeddings of local samples:

\begin{equation}
    \mathbf{p}_{q,c}^t = \frac{1}{|\mathcal{D}_{q,c}|} \sum_{(x_i,y_i)\in \mathcal{D}_{q,c}} f_{\theta_q}(x_i)
    \label{eq:local_proto}
\end{equation}

\noindent where $\mathcal{D}_{q,c} = \{(x_i, y_i) \in \mathcal{D}_q: y_i = c\}$ denotes the local samples of class $c$ at client $q$. These local prototypes are sent to the server, which aggregates them class-wise over the subset of clients that observed each class:

\begin{equation}
    \bar{\mathbf{p}}_c^t = \frac{1}{|\mathcal{Q}_c|} \sum_{q \in \mathcal{Q}_c} \mathbf{p}_{q,c}^{t}
    \label{eq:global_proto}
\end{equation}

\noindent where $\mathcal{Q}_c = \{q \in \mathcal{Q}: c \in \mathcal{C}_q\}$. This class-wise aggregation from partial observations is a defining property of prototype-based FL and gives rise to the noise characteristics analysed in Section~\ref{subsec:maturity_gap}. The total communication cost per round is $Q \cdot \mu \cdot d$ parameters, being $\mu$ the average number of classes per client, thus independent of model size and far below full-model  averaging methods such as FedAvg.

During local optimisation, these global prototypes act as representation targets that regularise the embedding space toward cross-client semantic consistency. The workflow at round $t$ proceeds as follows. The server broadcasts the current global prototype bank $\{\bar{\mathbf{p}}_c^{t-1}\}_{c \in \mathcal{C}}$ to all clients. Each client $q$ performs local training on $\mathcal{D}_q$, guided by the received prototypes as class-level representation anchors. After local training, client $q$ computes updated local prototypes via Eq.~\ref{eq:local_proto} and transmits $\{\mathbf{p}_{q,c}^{t}\}_{c \in \mathcal{C}_q}$ back to the server. The server updates the global bank via Eq.~\ref{eq:global_proto} and the next round begins. 

The objective is not only to separate classes locally but to shape the latent space so that representations of the same class concentrate around a shared semantic region across clients. By pulling representations toward their corresponding global prototype and enforcing separation from prototypes of other classes, clients learn representations that are simultaneously discriminative locally and consistent globally, despite never sharing raw data or model parameters.

\subsection{Representation Learning Objective}

Local training balances three objectives: discriminative classification, cross-client representation alignment, and inter-class separation. The composite loss at round $t$ is:

\begin{equation}
\mathcal{L} = \mathcal{L}_{\mathrm{CE}} + \lambda_t \, \mathcal{L}_{\mathrm{Proto}}+ \mathcal{L}_{\mathrm{proxy}},
\end{equation}

\noindent where $\lambda_t$ is a time-dependent weight term, whose scheduling is detailed in Section~\ref{sec:curriculum}. The three terms are defined as described below. 

The cross-entropy loss $\mathcal{L}_{\mathrm{CE}}$ optimises local classification over the observed classes $\mathcal{C}_q$ at client $q$: 

\begin{equation}
    \mathcal{L}_{\mathrm{CE}} = -\sum_{c \in \mathcal{C}_q} y_c\log g_{\phi_q}(f_{\theta q}(x_i)),
\end{equation}

The prototype alignment loss $\mathcal{L}_{\mathrm{Proto}}$ enforces cross-client representational consistency by pulling local embeddings towards their corresponding global prototype:

\begin{equation}
    \mathcal{L}_{\mathrm{Proto}} = \| f_{\theta_q}(x_i) - \bar{\mathbf{p}}_{c}^t \|_2^2, \quad c = y_i
\end{equation}

This MSE formulation induces gradients proportional to the embedding-prototype distance, a property that is central to the analysis in Section~\ref{subsec:maturity_gap}. 

The proxy regularisation loss $\mathcal{L}_{\mathrm{Proxy}}$ enforces inter-class separation on the hypersphere and is detailed in Section~\ref{subsec:proxy}

\subsection{Motivation: The Alignment-Maturity Gap}
\label{subsec:maturity_gap}

A key instability in prototype-based FL arises from alignment timing: global prototypes are imposed as fixed anchors from early rounds, before local representations or prototypes become reliable. The gradient of $\mathcal{L}_{\mathrm{Proto}}$ with respect to the feature extractor satisfies:

\begin{equation}
    \nabla_\theta \mathcal{L}_{\mathrm{Proto}}
    \propto
    \big( f_\theta(x) - \mathbf{p}_y \big)
    \frac{\partial f_\theta}{\partial \theta}.
    \label{eq:dilemma}
\end{equation}

\noindent This expression simply reflects the standard gradient structure of an MSE loss and shows that update magnitude depends on the embedding-prototype distance. In early rounds, this distance is large for two compounding reasons. First, local feature extractors $f_\theta$ have not yet organised a discriminative embedding space, producing high-variance, semantically uninformative representations. Second, global prototypes $\mathbf{p}_y$ are aggregated from these same unrefined embeddings, making them unreliable semantic anchors. The result is a self-reinforcing instability --the larger the embedding-prototype gap, the stronger the gradient that attempts to close it, and the more severely local feature learning is disrupted. We use the term \emph{alignment-maturity gap} to describe this interaction between the round at which global alignment is enforced and the round at which representations are stable enough to benefit from it.

\begin{figure}[t]
    \centering
    \includegraphics[width=\linewidth]{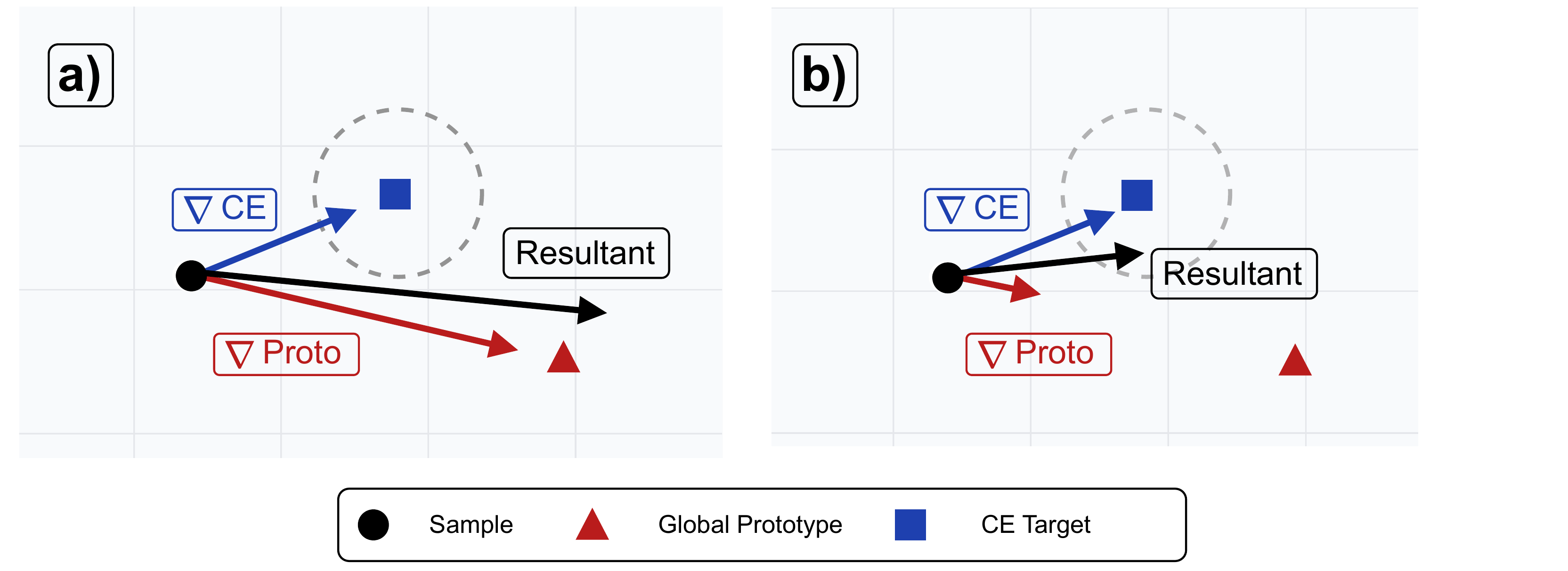}
    \caption{Conceptual illustration of the alignment-maturity gap: with a static alignment coefficient, early gradient domination suppresses local discriminative structure before it can consolidate.}
    \label{fig:double_penalty}
\end{figure}

This is related to, but distinct from, instability in proximal optimisation~\cite{li2020federated}, where aggressive regularisation toward an inconsistent global objective produces divergent updates. The distinction is the distance-dependent nature of the gradient in Eq.~\ref{eq:dilemma}: unlike a fixed proximal penalty, prototype alignment pressure intensifies precisely when the global reference is least trustworthy. Under a static alignment coefficient $\lambda$, this dynamic is uncontrolled (see Fig. \ref{fig:double_penalty}). This is a qualitative effect that has also been implicitly leveraged in prior warm-up and curriculum strategies in representation learning.

\subsection{Progressive Alignment Curriculum}
\label{sec:curriculum}

The alignment-maturity gap motivates controlling \emph{when} global alignment is introduced. FedSAP treats this as a curriculum problem: alignment pressure is introduced gradually, allowing local representations to develop discriminative structure before being pulled towards global consensus. Formally, the alignment weight $\lambda$ at round $t$ is:

\begin{equation}
    \lambda_t = M \cdot \min\left(\frac{t}{t_w},1\right)
    \label{eq:lambda}
\end{equation}

\noindent where $t_w$ is the curriculum length and $M$ is the maximum alignment coefficient. For $t < t_w$ the embedding space is shaped primarily by the local classification objective, with alignment pressure increasing linearly as representations stabilise. At $t = t_w$, the full coefficient $M$ is reached and held constant.

We adopt a linear schedule as the simplest monotone curriculum, avoiding the introduction of additional shape hyperparameters. Empirically, performance is stable across schedule shapes --cosine, sigmoid, and step-function variants produce comparable results (results in Appendix)-- consistent with the interpretation that the benefit derives from the principle of gradual introduction rather than any specific functional form. We therefore adopt the linear schedule for simplicity.

\subsection{Geometry-Driven Proxy Regularisation}
\label{subsec:proxy}

Prototype alignment alone encourages embeddings to concentrate near class centroids but does not explicitly enforce separation between classes. We address this through a proxy regularisation loss that structures the embedding space on a hypersphere, exploiting the global prototype bank already present at each client.

Rather than maintaining a separate set of learnable proxy parameters, we observe that the global prototype bank already provides class-conditional anchors in embedding space, making it a natural proxy set. We therefore construct the proxy bank directly from the server-distributed prototypes, avoiding any additional parameters or communication. This design choice couples the separation objective to the current state of global representations, requires no additional communication, and is grounded in the observation that well-formed prototypes are already distributed class-conditionally in embedding space, making them natural proxies for a cosine-softmax separation loss.

For a local batch $B$, embeddings and prototypes are normalised onto the unit hypersphere.

\begin{equation}
    \hat{h}_i = \frac{f_\theta(x_i)}{\|f_\theta(x_i)\|_2}, \qquad \hat{\mathbf{p}}_c = \frac{{\mathbf{p}}_c}{\|\mathbf{p}_c\|_2}
\end{equation}

Scaled cosine-similarity logits are then computed as $z_{i,c} = s \cdot \hat{h}_i^\top \hat{\mathbf{p}}_c$, where $s > 0$ is a temperature parameter that prevents gradient saturation on the hypersphere, following standard practice in hyperspherical metric learning. The proxy loss is a softmax cross-entropy over these logits:

\begin{equation}
    \mathcal{L}_{\mathrm{Proxy}} = - \frac{1}{|B|} \sum_{i \in B} \log \frac{\exp(z_{i, y_i})}{\sum_{c \in \mathcal{C}} \exp(z_{i,c})}
    \label{eq:proxy_loss}
\end{equation}

\subsection{Extension to Semi-Supervised Federated Learning}

An additional advantage of FedSAP is that it extends naturally to semi-supervised settings because its representation objectives do not fundamentally depend on full supervision. We adopt a two-phase protocol: a supervised phase of $T_{\mathrm{sup}}$ rounds establishes a stable prototype bank and a well-structured latent space; a subsequent semi-supervised phase leverages unlabelled client data to refine representations further.

At the start of the semi-supervised phase, each client $q$ has a local unlabelled pool of samples $\mathcal{U}_q$ of its known classes, maintaining the closed-set assumption. For each unlabelled sample $x_u \in \mathcal{U}_q$, a pseudo-label is assigned by the nearest cosine distance to the global prototype bank:

\begin{equation}
    \hat{c}_u = \arg\min_c \left(1 - \frac{\tilde{h}_u^\top \hat{\mathbf{p}}_c}{\|\tilde{h}_u\|_2 \|\hat{\mathbf{p}}_c\|_2}\right)
    \label{eq:pseudo_label}
\end{equation}

\noindent where $\tilde{h}_u = f_{\theta_q}(x_u)$ is computed under weak augmentation, following the weak-to-strong consistency paradigm of FixMatch~\cite{sohn2020fixmatch}. To reduce confirmation bias~\cite{arazo2020pseudo}, a sample is accepted into the active set $\mathcal{A}_q$ of pseudo-labelled samples only if the margin between the distances to its nearest prototype $p$ and its second-nearest prototype $p'$ satisfies:

\begin{equation}
    d(x_u, p') - d(x_u, p) > M \quad \text{and} \quad d(x_u, p) < D_{\max}
\end{equation}

\noindent Rather than manual tuning, $M$ and $D_{\max}$ are calibrated automatically at the end of the supervised phase from the distribution of margins and true-class distances observed on labelled data, making the thresholds adaptive to the geometry of each client's local embedding space.

For each accepted sample $x_u \in \mathcal{A}_q$, a strongly-augmented view produces the embedding $h_u^{(s)}$. A consistency loss penalises the cosine distance between this view and the target prototype:

\begin{equation}
    \mathcal{L}_{\mathrm{Cons}} = \frac{1}{|\mathcal{A}_q|} \sum_{x_u \in \mathcal{A}_q} \Bigl(1 - \hat{h}_u^{(s)\top} \hat{\mathbf{p}}_{\hat{y}_u} \Bigr)
\end{equation}

\noindent where $\hat{h}_{x_u}^{(s)}$ and $\hat{\mathbf{p}}_{\hat{c}_u}$ are $\ell_2$-normalised. This objective is geometrically consistent with $\mathcal{L}_{\mathrm{Proxy}}$: both operate on the unit hypersphere and pull representations toward class prototype anchors, meaning no architectural modification is required to accommodate unlabelled data. The full local objective is formulated as:

\begin{equation}
    \mathcal{L} = \mathcal{L}_{\mathrm{CE}} 
    + \lambda_t \mathcal{L}_{\mathrm{Proto}} 
    + \mathcal{L}_{\mathrm{Proxy}} 
    + \lambda_{\mathrm{Cons}} \mathcal{L}_{\mathrm{Cons}}
    \label{eq:full_loss_ssl}
\end{equation}

To propagate semi-supervised gains to the global prototype bank, clients return mixed prototypes that incorporate both labelled and accepted pseudo-labelled embeddings. For a given class $c$, the returned local prototype is computed as:
\begin{equation}
    \mathbf{p}_c^{\mathrm{local}} = 
    \mathrm{norm}\!\left(
    \frac{
        \sum_{i \in \mathcal{L}_c} h_i 
        + \alpha \sum_{j \in \mathcal{A}_q^c} h_j^{(s)}
    }{|\mathcal{L}_c| + \alpha|\mathcal{A}_q^c|}
    \right)
\end{equation}
where $\mathcal{A}_q^c$ is the subset of accepted active pseudo-labelled samples assigned to class $c$, and $\alpha$ controls the contribution of pseudo-labelled samples to the prototype estimate. This closing step ensures that representational improvements from the semi-supervised phase propagate back to the global prototype bank, benefiting all clients in subsequent rounds.

\section{Experimental Results}
\label{sec:results}

\subsection{Experimental Setup}

\paragraph{Datasets and Heterogeneity Protocol.} We evaluate FedSAP on three standard visual recognition benchmarks: FEMNIST~\cite{caldas2018leaf}, CIFAR-10, CIFAR-100~\cite{krizhevsky2009learning}. Following the protocol  established in FedProto~\cite{tan2022fedproto}, we adopt an $n$-way $k$-shot formulation to simulate statistical heterogeneity across clients. For each client, the number of available classes $n$ and samples per class $k$ are drawn from distributions parameterised by mean $(\mu_n, \mu_k)$ and standard deviation $\sigma$. To assess robustness across heterogeneity levels, we evaluate three partitioning configurations per dataset, ranging from moderate to severe non-IID conditions. Full dataset statistics and partition configurations are provided in the Appendix.

\paragraph{Architectures.} For FEMNIST, we use a CNN with two convolutional layers and two fully connected layers. For CIFAR-10 and CIFAR-100, we used a ResNet-18~\cite{he2015deepresiduallearningimage} initialised with ImageNet pre-trained weights.

\paragraph{Baselines.} We compare FedSAP against FedAvg~\cite{mcmahan2017communication} as the canonical parameter-aggregation baseline, and against two recent SOTA prototype-based FL methods, FedProto~\cite{tan2022fedproto}, and FedMPS~\cite{yang2025fedmps}, which are the most closely related to our formulation under heterogeneous client distributions.

\paragraph{Implementation Details.} 
All methods are implemented in PyTorch \footnote{code to be released upon acceptance}. Baselines follow their reference implementations, adapted to our data partitioning protocol. All experiments use 20 clients with full participation over 100 communication rounds, one local epoch per round, and a mini-batch size of 8. Optimisation uses SGD with learning rate 0.01 and momentum of 0.5. All results are reported as the mean $\pm$ standard deviation over 3 independent runs with seeds \{1234, 1235, 1236\}. Method-specific hyperparameters for baselines follow their respective publications. FedSAP hyperparameters are detailed in Appendix.

\paragraph{Evaluation Protocol.} All methods are evaluated on a centralised test set containing all classes. During evaluation, each client reports accuracy on classes present in its local training data. We evaluate two inference strategies: (i) classifier-head inference using the model classification layer, and (ii) prototype-based inference following the FedProto scheme, where samples are assigned to the nearest class prototype in the embedding space. We report top-1 accuracy and the number of parameters communicated per round.

\subsection{Performance Comparison}

\begin{table*}[t]

\small
\setlength{\tabcolsep}{3pt}
\resizebox{\textwidth}{!}{%
\begin{tabular}{l|ccc|ccc|ccc}
& \multicolumn{3}{c|}{\textbf{FEMNIST}} 
& \multicolumn{3}{c|}{\textbf{CIFAR-10}} 
& \multicolumn{3}{c}{\textbf{CIFAR-100}} \\
\cmidrule(lr){2-4} \cmidrule(lr){5-7} \cmidrule(lr){8-10}
\textbf{Method} 
  & $n{=}3$ & $n{=}4$ & $n{=}5$ 
  & $n{=}3$ & $n{=}4$ & $n{=}5$ 
  & $n{=}10$ & $n{=}20$ & $n{=}30$ \\
\midrule

\multicolumn{10}{c}{\textbf{Classifier-head inference}} \\
\midrule

FedAvg 
  & $33.60{\pm}15.12$ & $43.35{\pm}14.93$ & $47.26{\pm}18.05$ 
  & $89.32{\pm}0.79$  & $\mathbf{91.60{\pm}0.88}$  & $\mathbf{93.02{\pm}0.48}$ 
  & $71.34{\pm}0.55$  & $76.15{\pm}0.80$  & $78.07{\pm}0.75$ \\

FedProto 
  & $92.22{\pm}2.08$ & $90.81{\pm}1.46$ & $90.06{\pm}0.65$ 
  & $89.04{\pm}1.84$ & $86.36{\pm}0.75$ & $83.42{\pm}1.06$ 
  & $79.91{\pm}0.59$ & $73.04{\pm}0.97$ & $68.26{\pm}0.82$ \\

FedMPS 
  & $92.65{\pm}1.51$ & $\mathbf{91.10{\pm}1.93}$ & $\mathbf{90.15{\pm}0.39}$ 
  & $87.63{\pm}1.54$ & $84.08{\pm}0.68$ & $81.71{\pm}1.25$ 
  & $77.75{\pm}1.03$ & $72.41{\pm}1.13$ & $67.63{\pm}0.59$ \\

FedSAP 
  & $\mathbf{93.01{\pm}1.97}$ 
  & $90.71{\pm}1.76$ 
  & $89.58{\pm}0.58$ 
  & $\mathbf{92.68{\pm}1.29}$ 
  & $90.58{\pm}0.56$ 
  & $89.46{\pm}0.63$ 
  & $\mathbf{87.59{\pm}0.26}$ 
  & $\mathbf{82.55{\pm}0.75}$ 
  & $\mathbf{78.95{\pm}0.88}$ \\

\midrule

\multicolumn{10}{c}{\textbf{Prototype-based inference (nearest centroid)}} \\
\midrule

FedProto
  & $92.19{\pm}2.00$ & $90.68{\pm}1.43$ & $89.56{\pm}0.70$ 
  & $\mathbf{89.84{\pm}1.70}$ & $\mathbf{87.01{\pm}0.67}$ & $84.11{\pm}0.89$ 
  & $80.98{\pm}0.43$ & $73.65{\pm}0.88$ & $68.45{\pm}0.87$ \\

FedMPS
  & $92.67{\pm}1.52$ & $\mathbf{91.06{\pm}1.99}$ & $\mathbf{90.11{\pm}0.35}$ 
  & $88.11{\pm}1.60$ & $84.97{\pm}0.73$ & $82.87{\pm}1.42$ 
  & $79.70{\pm}0.64$ & $72.72{\pm}1.11$ & $65.97{\pm}0.54$ \\

FedSAP
  & $\mathbf{93.13{\pm}1.96}$ & $90.83{\pm}1.74$ & $89.46{\pm}0.45$ 
  & $86.98{\pm}2.65$ & $86.78{\pm}0.67$ & $\mathbf{87.00{\pm}0.38}$ 
  & $\mathbf{84.68{\pm}0.22}$ & $\mathbf{79.40{\pm}0.66}$ & $\mathbf{75.43{\pm}0.89}$ \\

\bottomrule
\end{tabular}%
}

\caption{Top-1 accuracy (\%, mean $\pm$ std, over 3 runs). Results are grouped by inference strategy. \textbf{Bold} indicates the best result within each column and inference group.}
\label{tab:main_results}
\end{table*}

\begin{figure*}[t]
    \centering
    \includegraphics[width=0.8\linewidth]{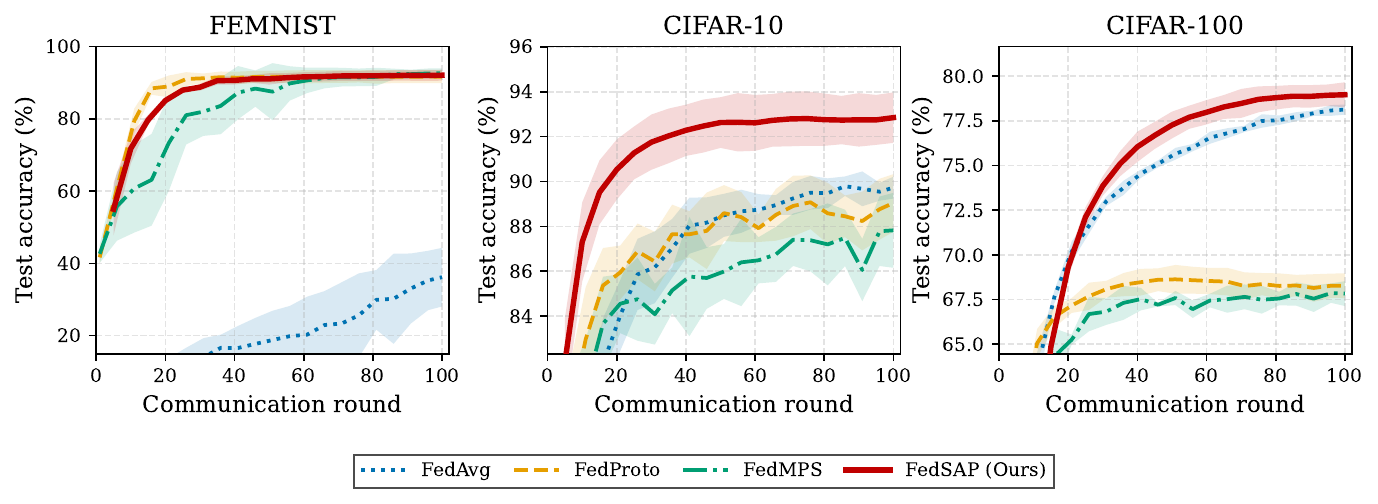}
    \caption{Accuracy convergence curves across different datasets. Performance evaluated every 5 rounds.}
    \label{fig:curves}
\end{figure*}

We evaluate across datasets, heterogeneity levels and inference strategies (Table~\ref{tab:main_results}, Figure~\ref{fig:curves}). FedSAP consistently outperforms prototype-based baselines under classifier-head inference on CIFAR-100, with gains increasing under higher heterogeneity, highlighting the impact of the alignment-maturity gap in noisy prototype regimes. The trend is consistent across initialisation regimes, whether training from scratch (FEMNIST) or starting from a pre-trained ResNet-18 (CIFAR-10/100), indicating that FedSAP does not depend on a specific backbone. On FEMNIST, FedAvg is unstable under high heterogeneity, while prototype-based methods are more robust. FedSAP achieves the best or near-best accuracy across all three splits with lower variance. On CIFAR-10 and CIFAR-100, FedAvg benefits from strong pre-training but requires substantially higher communication cost (over $350\times$ compared to prototype-based methods), making it less practical under equal communication budgets. Under these constraints, FedSAP consistently improves over FedProto and FedMPS, particularly as heterogeneity increases on CIFAR-100. This suggests that the progressive alignment curriculum is effective when the model already provide a good initialisation. Across inference strategies, FedSAP performs best with classifier-head inference, while baselines are largely insensitive to the choice. FedSAP sometimes underperforms in prototype-based inference, likely due to the cosine-softmax proxy objective aligning embeddings with classifier decision boundaries rather than Euclidean centroid proximity required by nearest-prototype decoding.

\paragraph{Latent Space Visualization.}

\begin{figure*}[t]
    
    \begin{subfigure}[t]{0.23\textwidth}
        \centering
        \includegraphics[width=\linewidth]{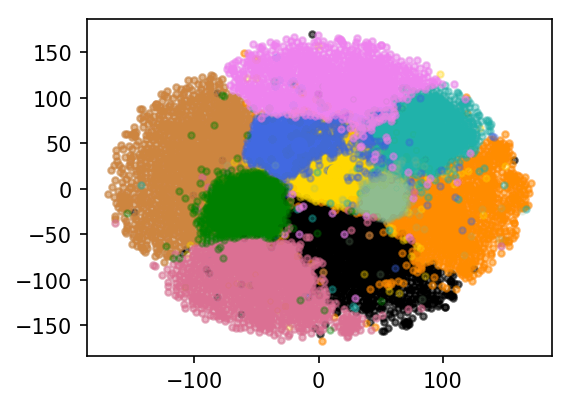}
        \caption{FedAvg}
    \end{subfigure}
    \hfill
    \begin{subfigure}[t]{0.23\textwidth}
        \centering
        \includegraphics[width=\linewidth]{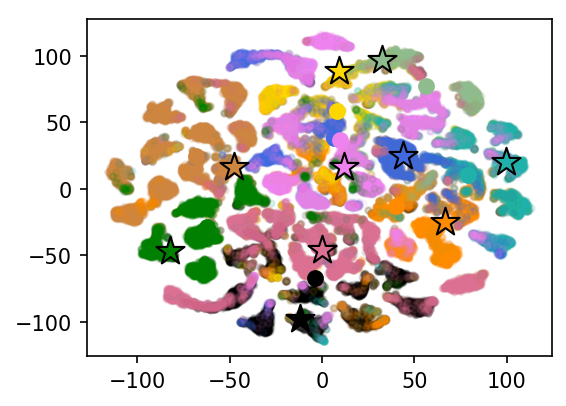}
        \caption{FedMPS}
    \end{subfigure}
    \hfill
    \begin{subfigure}[t]{0.23\textwidth}
        \centering
        \includegraphics[width=\linewidth]{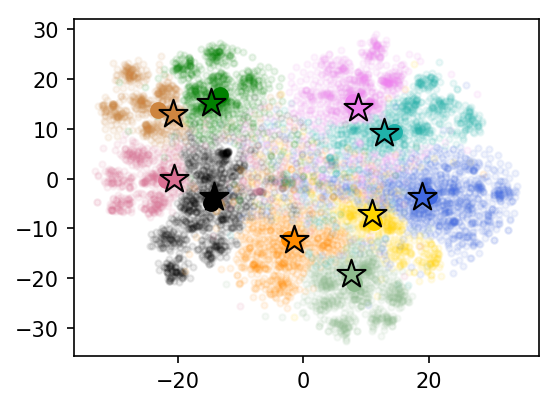}
        \caption{FedProto}
    \end{subfigure}
    \hfill
    \begin{subfigure}[t]{0.23\textwidth}
        \centering
        \includegraphics[width=\linewidth]{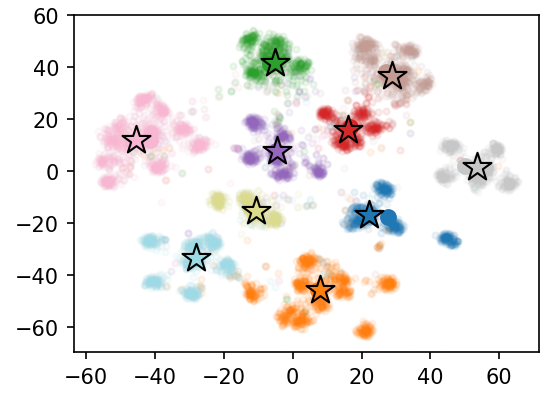}
        \caption{FedSAP (Ours)}
    \end{subfigure}

    \caption{t-SNE visualisation of the embeddings and prototypes (shown as stars) produced by the selected FL methods across CIFAR-10 dataset. Note that for FedAvg, there are no global prototypes since they are not generated.}
    \label{fig:ours_evolution}
\end{figure*}

The t-SNE visualisations in Figure~\ref{fig:ours_evolution} provide qualitative evidence for the representational advantage of FedSAP. Qualitatively, FedAvg produces diffuse, poorly separated clusters; FedProto and FedMPS achieve moderate compactness; and FedSAP produces tighter intra-class clusters with cleaner inter-class boundaries. Global prototypes (shown as stars) are well-centred within their respective clusters, indicating that the proxy regularisation has structured the embedding space geometrically rather than merely pulling embeddings toward noisy anchors. This confirms that the gains in Table~\ref{tab:main_results} reflect a genuinely better-organised representation space, not simply better-calibrated decision boundaries.

\paragraph{Communication Efficiency.}

\begin{table}[t]
\centering
\small
\begin{tabular}{l l r r r}
\toprule
\textbf{Method} & \textbf{Cost per round} & \multicolumn{3}{c}{\textbf{\#Comm. Params\textsuperscript{†}}} \\
\cmidrule(lr){3-5}
& & \textbf{FEMNIST} & \textbf{C10} & \textbf{C100} \\
\midrule
FedAvg    
  & $|\theta|$ (full model) 
  & $849$ & $11{,}182$ & $11{,}182$ \\
FedMPS    
  & $|\theta_s| + N \cdot \mu_n \cdot d$ 
  & $78$ & $43$ & $241$ \\
FedProto  
  & $N \cdot \mu_n \cdot d$ 
  & $\mathbf{3}$ & $\mathbf{31}$ & $\mathbf{102}$ \\
\midrule
\textbf{FedSAP}    
  & $N \cdot \mu_n \cdot d$ 
  & $\mathbf{3}$ & $\mathbf{31}$ & $\mathbf{102}$ \\
\bottomrule
\end{tabular}
\caption{Communication cost per round. $N$: number of clients; $\mu_n$: mean classes per client; $d$: embedding dimension; $|\theta|$: full model parameters; $|\theta_s|$: shared sub-model in FedMPS. \textsuperscript{†}Representative values ($\times 10^3$ params) for FEMNIST and CIFAR-10 with $n=3$, and CIFAR-100 with $n=10$. FedSAP matches FedProto exactly.}
\label{tab:comm_cost}
\end{table}

Table~\ref{tab:comm_cost} confirms that FedSAP preserves the minimal communication footprint of FedProto across all settings, exchanging only class-level prototype vectors. FedMPS incurs substantially higher costs due to its shared sub-model components, and FedAvg communicates full model parameters --several orders of magnitude more than prototype-based methods. FedSAP therefore achieves 
its representational improvements at no additional communication cost.

\paragraph{Robustness to Data Scarcity.}

\begin{figure}[t]
    \centering
    \includegraphics[width=0.8\linewidth]{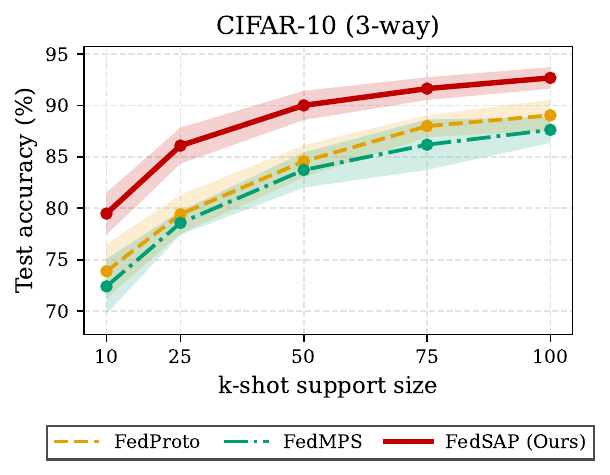}
    \caption{CIFAR-10 3-way test accuracy with classifier inference vs.\ $k$-shot size.}
    \label{fig:data_heterogeneity}
\end{figure}

Figure~\ref{fig:data_heterogeneity} explores the sensitivity to the number of local samples per class $k$ on CIFAR-10. FedSAP maintains a consistent margin over FedProto and FedMPS across all values of $k$, including the most data-scarce regime where only a handful of samples per class are available locally. The relative gap is preserved across the full range tested, suggesting that the geometry-driven proxy regularisation provides useful class separation signal even when local prototype estimates are unreliable due to sample sparsity.

\subsection{Semi-Supervised Extension}

\begin{table}[t]
\centering
\small
\setlength{\tabcolsep}{4pt}
\begin{tabular}{ll|cc|cc}
\toprule
& & \multicolumn{2}{c|}{\textbf{Supervised}}
& \multicolumn{2}{c}{\textbf{+ SSL Refinement}} \\
\cmidrule(lr){3-4}\cmidrule(lr){5-6}
\textbf{Dataset} & \textbf{Setting}
& Sil. & Acc.
& Sil. & Acc. \\
\midrule
\multirow{2}{*}{FEMNIST}
  & $|U|{=}600$  & 0.40 & 87.34 & \textbf{0.67} & \textbf{89.87} \\
  & $|U|{=}1000$ & 0.40 & 87.34 & \textbf{0.67} & \textbf{89.77} \\
\midrule
CIFAR-10 & $k{=}25$ & 0.14 & 69.18 & \textbf{0.24} & \textbf{73.98} \\
CIFAR-10 & $k{=}60$ & 0.23 & 81.52 & \textbf{0.34} & \textbf{84.61} \\
\bottomrule
\end{tabular}

\caption{Effect of the semi-supervised refinement stage on silhouette score and top-1 accuracy. The supervised phase runs for 100 rounds; refinement adds 50 further rounds using unlabelled client data. FEMNIST uses $n{=}4$, $k{=}60$, varying the unlabelled pool size $|U|$. CIFAR-10 uses $n{=}3$, $|U|{=}600$, with $k{=}25$ and $k{=}60$ labelled shots per class to simulate data scarcity.}
\label{tab:ssl_refinement}
\end{table}

We treat the semi-supervised extension as a preliminary exploration of representation refinement rather than a fully formalised SSL protocol, and evaluate it on FEMNIST and CIFAR-10 using the same model checkpoints produced by supervised training as the starting point, with no architectural changes required. Table~\ref{tab:ssl_refinement} reports the silhouette score and accuracy before and after 50 rounds of unlabelled refinement. Across both datasets, the semi-supervised phase improves both representation quality and distance-based classification accuracy, with the largest gains observed on CIFAR-10 under data scarcity ($k{=}25$), where the supervised phase produces a relatively weak embedding space that unlabelled consistency training can meaningfully refine. On FEMNIST, the unlabelled pool size has little effect on final accuracy, suggesting the limiting factor is the quality of pseudo-labels rather than their quantity.

This also highlights a limitation of the closed-set setting, where unlabelled data is assumed to come from known classes; relaxing this to open-set  distributions is left for future work. We omit dedicated semi-supervised federated baselines, as the aim is not protocol comparison but isolating whether unlabelled refinement reorganises learned representation.

\begin{figure*}[t]
    \centering

    \begin{subfigure}[b]{0.33\textwidth}
        
        \includegraphics[width=\linewidth]{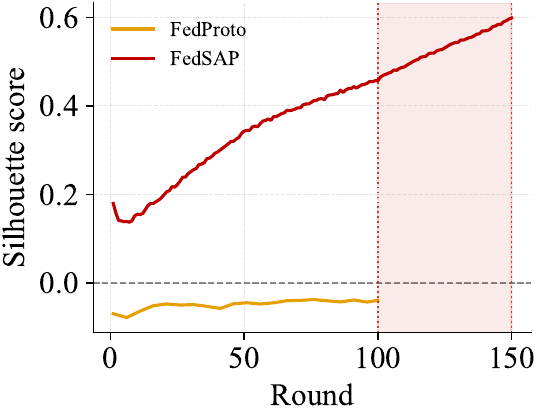}
        \caption{Silhouette score evolution}
        \label{fig:silhouette_ssl}
        
    \end{subfigure}
    \hfill
    \begin{subfigure}[b]{0.30\textwidth}
        
        \includegraphics[width=\linewidth]{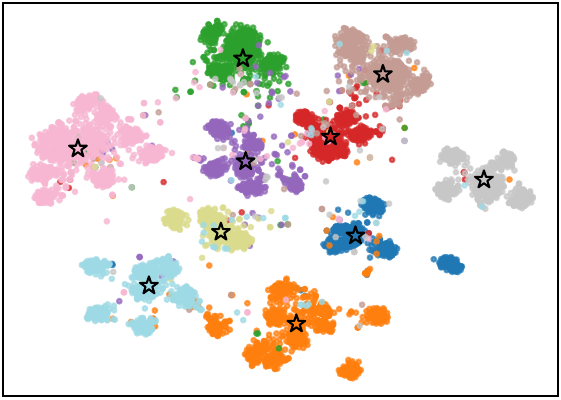}
        \caption{Supervised Phase}
        \label{fig:tsne_100}
        
    \end{subfigure}
    \hfill
    \begin{subfigure}[b]{0.30\textwidth}
        
        \includegraphics[width=\linewidth]{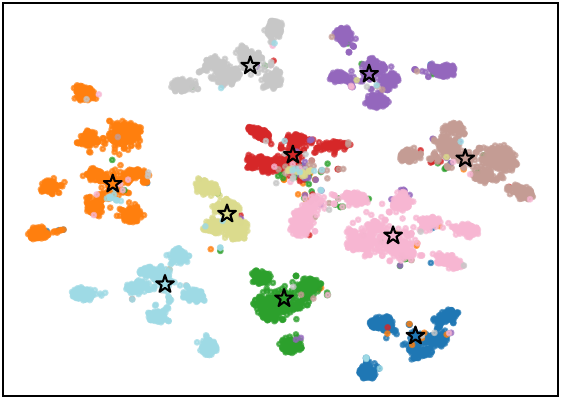}
        \caption{Semi-supervised Phase}
        \label{fig:tsne_150}
        
    \end{subfigure}

    \caption{
    (a) Evolution of the Silhouette score across supervised and semi-supervised phases, comparing FedProto and FedSAP on CIFAR-10 with $n=3$.
    (b--c) t-SNE projections of local client embeddings, where stars represent global class prototypes.
    }
    \label{fig:representation_analysis}
\end{figure*}

Figure~\ref{fig:representation_analysis} provides further evidence of this dynamic. The silhouette trajectory (Fig.~\ref{fig:silhouette_ssl}) shows that FedProto's embedding space remains poorly structured throughout training, with scores hovering near zero, whereas FedSAP progressively organises class structure during the supervised phase and continues improving in the semi-supervised phase, reaching substantially higher silhouette values by round 150. The t-SNE projections (Figs.~\ref{fig:tsne_100} and~\ref{fig:tsne_150}) confirm this evolution: by the end of the supervised phase the space is already well partitioned, with prototypes centred within class clusters, while by the end of training intra-class variance is further reduced and inter-class margins increase, indicating that consistency training refines rather than disrupts the learned representation structure.
\subsection{Ablation Study} 

\begin{table}[t]
\centering
\small
\begin{tabular}{cc cc}
\toprule
\textbf{Curriculum} 
& \textbf{Proxy} 
& \textbf{Proto Acc.} 
& \textbf{Head Acc.} \\
\midrule
-- & -- & 79.70 & 79.91 \\
\checkmark & -- & 82.52 & 80.85 \\
-- & \checkmark & 82.81 & 87.45 \\
\checkmark & \checkmark & \textbf{84.68} & \textbf{87.59} \\
\bottomrule
\end{tabular}

\caption{Ablation on CIFAR-100 ($n{=}10$). 
Proto Acc.: nearest-centroid inference.
Head Acc.: classifier-head inference.}
\label{tab:ablation}
\end{table}

Table~\ref{tab:ablation} isolates the contribution of each FedSAP component on CIFAR-100 ($n{=}10$). The progressive alignment curriculum alone improves upon FedProto by approximately 1pp under classifier-head inference, directly validating the alignment-maturity gap hypothesis: delaying full alignment pressure provides a more stable foundation for local feature learning. The proxy regularisation alone accounts for the majority of the total gain, lifting head accuracy by over 7pp relative to the baseline, which confirms that geometry-driven inter-class separation is the primary driver of representation quality. Combining both components further improves prototype-based inference accuracy, as the curriculum stabilises the embedding space during early rounds when proxy gradients are most sensitive to prototype noise. Sensitivity analyses for the curriculum schedule shape, warm-up onset, and proxy temperature are provided in Supplementary Material: performance is stable across a wide range of settings.

\section{Conclusion}
\label{sec:conclusions}
We addressed the alignment-maturity gap in prototype-based FL, where global alignment is imposed before local representations are mature enough to benefit from it. FedSAP mitigates this through a progressive alignment curriculum and geometry-aware proxy regularisation on a shared prototype bank, promoting inter-class separation on the hypersphere without modifying the communication protocol. Across three benchmarks, this yields consistent improvements over prototype-based baselines, which become more pronounced under higher heterogeneity, and generalises across initialisation regimes. The proxy-structured embedding space is geometrically compatible with downstream tasks such as retrieval and few-shot recognition, where hyperspherical representations are known to transfer effectively~\cite{kim2020proxy}.

FedSAP extends to semi-supervised settings with minimal modification, suggesting that scheduled alignment is a general design principle rather than protocol-specific fix. While prototype-based semi-supervised FL has received limited attention, methods such as \cite{kim2022protofssl} and \cite{pan2025fedppl} jointly optimise labelled and unlabelled objectives from early rounds, leaving the cold-start issue of unreliable early representations unaddressed. We view this extension as an initial step. Future work could explore open-set scenarios with unseen classes via out-of-distribution detection~\cite{jeong2020federated}, or incorporating local self-supervised pre-training~\cite{zhuang2021collaborative} to reduce reduce reliance on scheduling. A more extensive evaluation on retrieval and few-shot benchmarks would further clarify how far the learned geometry generalises beyond classification.

\bibliographystyle{plain}
\bibliography{egbib}

\appendix

\section{Dataset Configurations}
\label{app:datasets}

\paragraph{Datasets}
We evaluate on three standard federated learning benchmarks. \textbf{FEMNIST}~\cite{caldas2018leaf} is a character recognition dataset derived from the EMNIST benchmark, naturally partitioned by writer identity and well-suited to heterogeneous FL evaluation. \textbf{CIFAR-10} and \textbf{CIFAR-100}~\cite{krizhevsky2009learning} are standard object recognition benchmarks with 10 and 100 classes respectively, partitioned synthetically according to the $n$-way $k$-shot protocol described below.

\paragraph{Heterogeneity Protocol} Following~\cite{tan2022fedproto}, we adopt an $n$-way $k$-shot formulation in which each client receives data from $n \sim \mathcal{N}(\mu_n, \sigma^2)$ classes with $k \sim \mathcal{N}(\mu_k, \sigma^2)$ samples per class. The parameters $(\mu_n, \mu_k, \sigma)$ are varied to produce three heterogeneity levels per dataset. Lower $\mu_n$ corresponds to higher heterogeneity: each client observes fewer classes, increasing distributional divergence across clients.

\begin{table}[h]
\begin{center}
\begin{tabular}{l|ccc}
\toprule
\textbf{Dataset} & $\mu_n$ & $\mu_k$ & $\sigma$ \\ \midrule
\multirow{3}{*}{FEMNIST} & 3 & 100 & 1 \\
 & 4 & 100 & 1 \\
& 5 & 100 & 2 \\ \midrule
\multirow{3}{*}{CIFAR-10} & 1 & 100 & 1 \\
 & 4 & 100 & 1 \\
 & 5 & 100 & 2 \\ \midrule
\multirow{3}{*}{CIFAR-100} & 10 & 50 & 4 \\
 & 20 & 50 & 6 \\
 & 30 & 50 & 8 \\ 
\bottomrule
\end{tabular}
\end{center}
\caption{$n$-way $k$-shot configurations per dataset and heterogeneity level. $\mu_n$ controls the expected number of classes per client; lower values correspond to more severe distributional divergence. Note that for CIFAR-100, severe heterogeneity still exposed clients to more classes than for CIFAR-10 due to larger label space.}
\label{tab:data_configs}
\end{table}

\paragraph{Implementation details.}
We train all methods using 20 clients for a total of 100 communication rounds. Each client performs 1 local epoch per round with a batch size of 8. Optimisation is performed using SGD with learning rate 0.01 and momentum 0.5.

We use a semi-supervised setting where unlabelled data is introduced during the first 50 communication rounds, with a per-client unlabelled pool of 600 samples. The proxy logits are scaled by a factor of 32. The prototype loss is introduced after round 20 and linearly increases from 0 to 0.7 until round 100.

\subsection{Ablation on the prototype-loss schedule shape.}

\begin{figure}[t]
    \centering
    \includegraphics[width=0.7\linewidth]{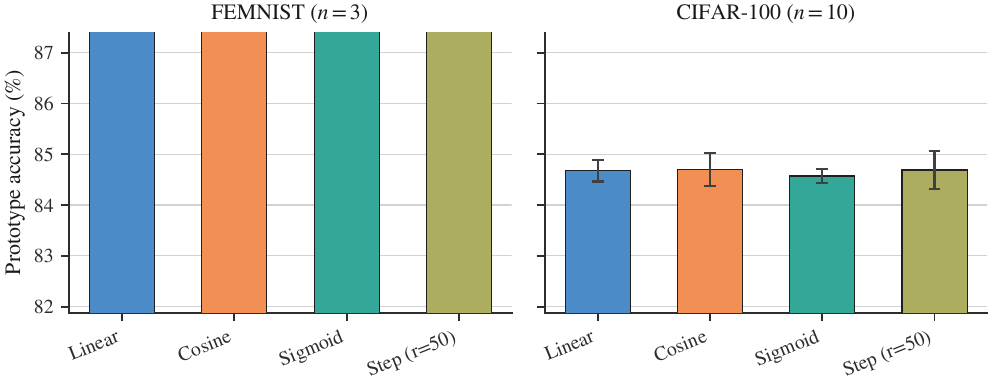}
    \caption{
    Comparison of prototype-loss weight schedules on FEMNIST ($n{=}3$, left) and CIFAR-100 ($n{=}10$, right). Results report prototype-based inference accuracy (nearest prototype). Smooth ramp schedules yield consistently strong performance across datasets.
    }
    \label{fig:appendix_exp_a_bars}
\end{figure}

We analyse how the temporal evolution of the prototype alignment weight $\lambda_{\mathrm{proto}}$ influences the learned federated representation. All configurations use the same training setup and differ only in the scheduling rule applied to $\lambda_{\mathrm{proto}}$ throughout training. Figure~\ref{fig:appendix_exp_a_bars} compares five representative schedules on FEMNIST and CIFAR-100.

Overall, smooth ramp strategies (linear, cosine, and sigmoid) achieve similar and consistently strong performance, indicating that the precise shape of the ramp is less important than progressively increasing prototype supervision during later training stages. In contrast, the step schedule represents a more abrupt allocation strategy, introducing it too suddenly, producing greater variability. These results support the use of a gradual ramped schedule as a stable and robust design choice.

\subsection{Effect of the prototype schedule start round}

\begin{figure}[t]
    \centering
    \includegraphics[width=0.9\linewidth]{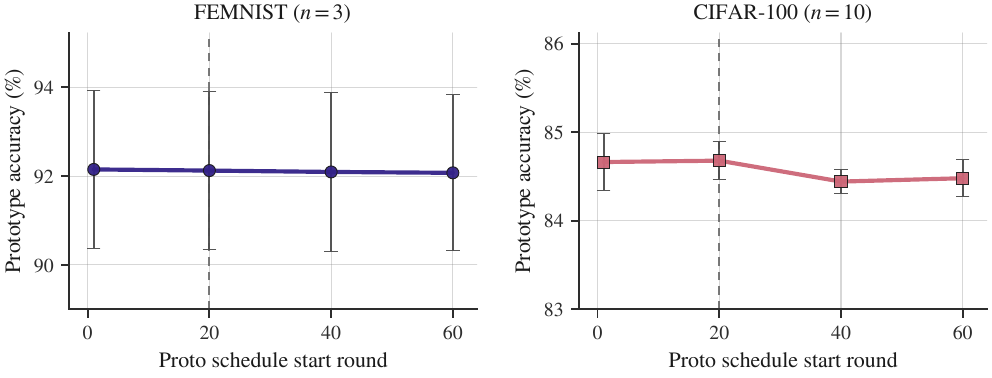}
    \caption{
    Sensitivity of prototype-based inference accuracy to the start round of the linear prototype-weight ramp on FEMNIST ($n{=}3$, left) and CIFAR-100 ($n{=}10$, right). For each configuration, $\lambda_{\mathrm{proto}}$ increases linearly from $r_{\mathrm{start}}$ until round $100$.
    }
    \label{fig:appendix_exp_b_start}
\end{figure}

We further study the effect of delaying prototype supervision by varying the start round $r_{\mathrm{start}}$ of the linear ramp schedule. Before $r_{\mathrm{start}}$, training is driven only by the classification objective, allowing the backbone to first learn discriminative local representations before prototype alignment becomes active.

As shown in Figure~\ref{fig:appendix_exp_b_start}, both very early and very late activation of prototype supervision tend to reduce performance. Starting from the first round can over-constrain the representation before stable class structure emerges, whereas activating alignment too late leaves insufficient time for prototype consolidation. Intermediate start rounds achieve the best trade-off, with the default choice $r_{\mathrm{start}}{=}20$ providing consistently strong and stable results across datasets.

\subsection{Effect of the proxy scaling factor}

\begin{figure}[t]
    \begin{center}
    \includegraphics[width=0.85\linewidth]{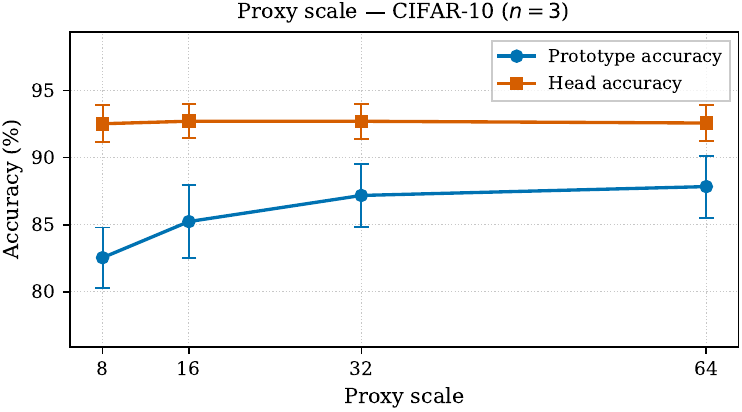}
    \caption{
    Ablation study on the proxy scale parameter $s$ on CIFAR-10 with $n{=}3$. We report both classifier-head inference accuracy and prototype-based inference accuracy (nearest prototype). Error bars denote standard deviation across three runs.
    }
    \label{fig:scale_ablation}
    \end{center}
\end{figure}

We perform an ablation study on the hyperparameter $s$ (scale proxy), which modulates the temperature of the cosine-similarity logits in the proxy regularisation objective ($\mathcal{L}_{\mathrm{Proxy}}$). We evaluate values of $s \in \{8,16,32,64\}$ to assess the impact of logit scaling on the hyperspherical embedding space. Figure~\ref{fig:scale_ablation} reports results on CIFAR-10 with $n{=}3$ for both classifier-head inference and prototype-based inference. Lower scaling factors ($s{=}8$ and $s{=}16$) lead to substantially worse performance, with accuracies remaining under $85\%$, indicating insufficient separation between class representations in the embedding space. In contrast, larger values ($s{=}32$ and $s{=}64$) produce consistently stronger and nearly identical performance across both inference strategies. Based on these results, we select $s{=}32$ as the default configuration for all experiments.

\subsection{Visual Latent Space Progression}
\label{appendix:tsne_trajectory}

\begin{figure*}[t]
    \centering
    \includegraphics[width=1\textwidth]{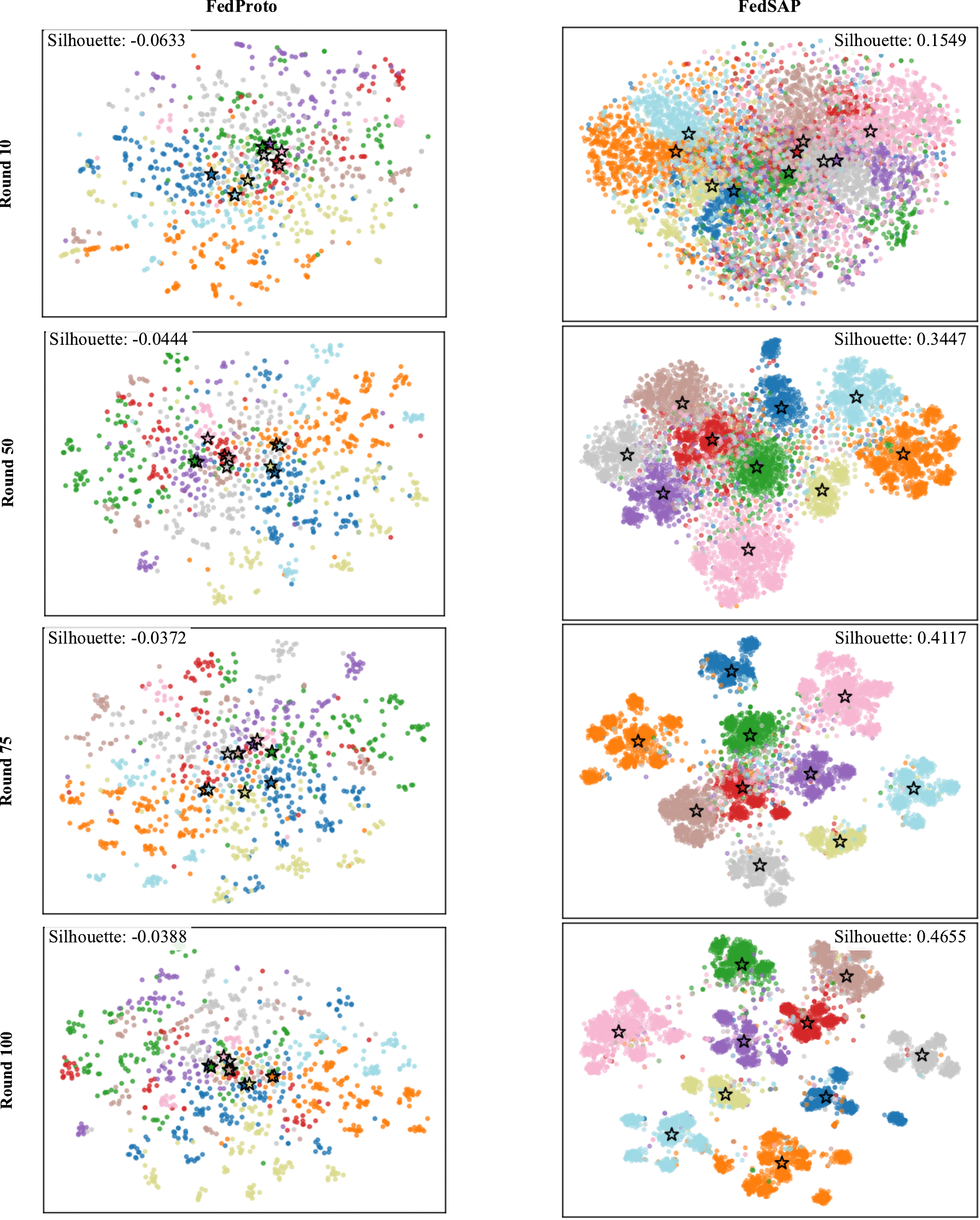}
    \caption{Visual trajectory of local client embeddings at Rounds 10, 50, 75, and 100, comparing FedProto (left) and FedSAP (right). Black stars represent global class prototypes. FedSAP progressively projects and separates classes into tight clusters on the unit hypersphere, while FedProto's representations remain congested.}
    \label{fig:tsne_rounds_comparison}
\end{figure*}

To illustrate how FedSAP structures the latent space over time, Figure~\ref{fig:tsne_rounds_comparison} compares the t-SNE projections of local client representations against FedProto at Rounds 10, 50, 75, and 100 of the supervised phase.

The visual trajectory reveals a stark contrast in the quality of the representation. Under FedProto (left column), the embedding space fails to organize, as evidenced by a consistently negative Silhouette score (ranging from $-0.0633$ to $-0.0388$). Throughout training, class representations remain heavily mixed, and the global prototypes (black stars) remain congested in a central clump, unable to guide local features. Conversely, FedSAP (right column) prevents early-round representation collapse. Starting with a Silhouette score of $0.1549$ at Round 10, the global prototypes are quickly dispersed. By Round 50, distinct boundaries and inter-class margins emerge, culminating in highly compact, isolated clusters centered on their global prototype anchors by Round 100 ($\text{Silhouette} = 0.4655$).

\end{document}